\documentclass[conference]{IEEEtran}

\usepackage[T1]{fontenc}

\IEEEoverridecommandlockouts
\usepackage{cite}
\usepackage{amsmath,amssymb,amsfonts}
\usepackage{algorithmic}
\usepackage{graphicx}
\usepackage{textcomp}
\usepackage{xcolor}
\usepackage{tabularx}
\usepackage{multirow}
\usepackage{booktabs}
\usepackage{multirow}
\usepackage[nolist]{acronym}
\def\BibTeX{{\rm B\kern-.05em{\sc \kern-.025em b}\kern-.08em
    T\kern-.1667em\lower.7ex\hbox{E}\kern-.125emX}}

\usepackage{xspace}
\usepackage{textcomp}
\newcommand{\microCaspian}{\textmu Caspian\xspace}

\begin{document}

% \author{\IEEEauthorblockN{Anonymous Author(s)}}

\author{\IEEEauthorblockN{James Ghawaly Jr.}
\IEEEauthorblockA{\textit{Computer Science and Engineering} \\
\textit{Louisiana State University}\\
Baton Rouge, United States \\
jghawaly@lsu.edu}
\and
\IEEEauthorblockN{Andrew Nicholson}
\IEEEauthorblockA{\textit{Physics} \\
\textit{Oak Ridge National Laboratory}\\
Oak Ridge, United States \\
nicholsonad@ornl.gov}
\and
\IEEEauthorblockN{Catherine Schuman}
\IEEEauthorblockA{\textit{Electrical Engineering and Computer Science} \\
\textit{The University of Tennessee Knoxville}\\
Knoxville, United States \\
cschuman@utk.edu}
\and
\IEEEauthorblockN{Dalton Diez}
\IEEEauthorblockA{\textit{Computer Science and Engineering} \\
\textit{Louisiana State University}\\
Baton Rouge, United States \\
ddiez2@lsu.edu}
\and
\IEEEauthorblockN{Aaron Young}
\IEEEauthorblockA{\textit{Computer Science and Mathematics} \\
\textit{Oak Ridge National Laboratory}\\
Oak Ridge, United States \\
youngar@ornl.gov}
\and
\IEEEauthorblockN{Brett Witherspoon}
\IEEEauthorblockA{\textit{Electrification and Energy Infrastructures} \\
\textit{Oak Ridge National Laboratory}\\
Oak Ridge, United States \\
witherspoocb@ornl.gov}}

\title{Exploring Spiking Neural Networks for Binary Classification in Multivariate Time Series at the Edge
\thanks{This research was supported by the U.S. National Nuclear Security Administration (NNSA) Office of Defense Nuclear Nonproliferation Research and Development within the U.S. Department of Energy under Contract AC05-00OR22725.}
}
% \thanks{This research was supported by the U.S. National Nuclear Security Administration (NNSA) Office of Defense Nuclear Nonproliferation Research and Development within the U.S. Department of Energy under Contract AC05-00OR22725.}

\maketitle
% copyright---------------------
% \IEEEpubid{\begin{minipage}{\linewidth}
% \centering\footnotesize
% © 2025 IEEE. Personal use of this material is permitted. Permission from IEEE must be obtained for all other uses.
% \end{minipage}}
\IEEEpubid{%
  \begin{minipage}{\columnwidth}
  \vspace{4\baselineskip}%
  \centering\footnotesize
  \copyright~2025 IEEE. Personal use of this material is permitted.
  Permission from IEEE must be obtained for all other uses.
\end{minipage}
}
% copyright---------------------
\begin{abstract}
% In this work, we investigate spiking neural networks (SNNs) as a low-power, high-performance solution for step-wise binary classification in multivariate time series. We introduce a hardware-software co-design approach that leverages the Evolutionary Optimization of Neuromorphic Systems (EONS) algorithm to evolve compact SNNs directly within a simulator of an FPGA-based neuromorphic platform, \microCaspian{}.

% We first conduct an in-depth study of detecting low signal-to-noise ratio radioactive sources in gamma-ray spectral data, using application-specific optimizations in data preprocessing, fitness function design, and spike encoding. For this application, our evolved SNNs, with as few as 49 neurons and 66 synapses, achieve a true positive rate (TPR) of 67.1\% at a false alarm rate of 1/hr. This outperforms both PCA-based (TPR = 42.7\%) and deep learning (TPR = 49.8\%) baselines under the same constraints. A hardware deployment on \microCaspian{} confirms inference at just 2~mW power consumption with 20.2~ms latency per time step.

% To assess broader applicability, we apply the same framework, without any application-specific modifications, to seizure detection in scalp EEG recordings. A majority-vote ensemble of three evolved SNNs achieves 95\% sensitivity and a 16\% false positive rate, comparable to recent deep learning approaches while remaining deployable at the edge.

We present a general framework for training spiking neural networks (SNNs) to perform binary classification on multivariate time series, with a focus on step-wise prediction and high precision at low false alarm rates. The approach uses the Evolutionary Optimization of Neuromorphic Systems (EONS) algorithm to evolve sparse, stateful SNNs by jointly optimizing their architectures and parameters. Inputs are encoded into spike trains, and predictions are made by thresholding a single output neuron's spike counts. We also incorporate simple voting ensemble methods to improve performance and robustness.

To evaluate the framework, we apply it with application-specific optimizations to the task of detecting low signal-to-noise ratio radioactive sources in gamma-ray spectral data. The resulting SNNs, with as few as 49 neurons and 66 synapses, achieve a 51.8\% true positive rate (TPR) at a false alarm rate of 1/hr, outperforming PCA (42.7\%) and deep learning (49.8\%) baselines. A three-model any-vote ensemble increases TPR to 67.1\% at the same false alarm rate. Hardware deployment on the \microCaspian neuromorphic platform demonstrates 2~mW power consumption and 20.2~ms inference latency.

We also demonstrate generalizability by applying the same framework, without domain-specific modification, to seizure detection in EEG recordings. An ensemble achieves 95\% TPR with a 16\% false positive rate, comparable to recent deep learning approaches with significant reduction in parameter count.

\end{abstract}


\begin{acronym}

\acro{AI}{Artificial Intelligence}
\acro{SNN}{Spiking Neural Network}
\acro{ANN}{Artificial Neural Network}
\acro{CNN}{Convolutional Neural Network}
\acro{LLM}{Large Language Model}
\acro{RNN}{Recurrent Neural Network}
\acro{ML}{Machine Learning}
\acro{DL}{Deep Learning}
\acro{EONS}{Evolutionary Optimization for Neuromorphic Systems}
\acro{LIF}{Leaky Integrate-and-Fire}
\acro{LSTM}{Long Short-term Memory}
\acroplural{LSTM}[LSTMs]{Long Short-term Memories}
\acro{GRU}{Gated Recurrent Unit}
\acro{EEG}{Electroencephalogram}
\acro{MCC}{Matthews Correlation Coefficient}
\acro{AUC}{Area Under the Receiver Operating Characteristic Curve}
\acro{ORNL}{Oak Ridge National Laboratory}
\acro{HEU}{Highly Enriched Uranium}
\acro{WGPU}{Weapons Grade Plutonium}
\acro{SNR}{Signal to Noise Ratio}
\acro{ROC}{Receiver Operating Characteristic}
\acro{TTAC}{Technical Testing and Analysis Center}
\acro{MCU}{Microcontroller Unit}
\acro{FPGA}{Field Programmable Gate Array}
\acro{SAD}{Spectral Anomaly Detector}
\acro{FAR}{False Alarm Rate}
\acro{PCA}{Principal Component Analysis}
\acro{ARAD}{Autoencoder Radiation Anomaly Detector}
\acro{NEP}{Neuromorphic Event Processor}
\acro{IF}{Integrate-and-Fire}
\acro{OESNN}{Online Evolving Spiking Neural Network}
\acro{FCNN}{Fully Convolutional Neural Network}
\acro{FPR}{False Positive Rate}
\acro{FP}{False Positive}
\acro{RC}{Reservoir Computing}
\acro{SLRC}{Spiking Legendre Reservoir Computing}
\acro{LSNN}{Legendre Spiking Neural Network}
\acro{SPI}{Serial Peripheral Interface}
\acro{DMA}{Direct Memory Access}

\end{acronym}

\section{Introduction}\label{sec:intro}
The rapid advancement of generative \ac{DL} technology has significantly transformed human interaction with information, enabling the automation of diverse tasks. However, this progress has highlighted critical limitations, including the substantial energy requirements of large-scale models \cite{strubell2020energy}, reliance on increasingly scarce and expansive datasets \cite{villalobos2024position}, and challenges in achieving robust reasoning and adaptability \cite{valmeekam2022large,mirzadeh2024gsm,jiang2024peek}. Neuromorphic computing, inspired by the proven efficiency and adaptability of biological neural systems, offers a potential pathway to address these challenges \cite{roy2019towards,schuman2022opportunities,calimera2013human,schuman2017survey}. By leveraging the demonstrated success of biological neural networks as a model for computation, neuromorphic systems aim to replicate their efficiency and versatility. Through hardware-software co-design, these systems integrate energy-efficient computational architectures with biologically plausible \ac{SNN} models, facilitating the development of scalable, sustainable, and adaptive \ac{AI} solutions. 

While \acp{SNN} currently do not outperform deep learning models in most machine learning applications, they offer significant advantages in resource-constrained scenarios, such as long-term edge deployments, where traditional deep learning approaches may be impractical. \acp{SNN} offer the promise of high performance while maintaining low power through event driven computing.  This offers significant advantages in resource-constrained scenarios, such as long-term edge deployments, where traditional deep learning approaches may be impractical. A compelling example is binary classification in multivariate time series data, a task critical to many real-world applications such as security monitoring and healthcare diagnostics. Detecting weak signals of interest in time series data requires models capable of identifying subtle, temporally distributed patterns within noisy and high-dimensional signals. At the edge, these challenges are compounded by the need for energy efficiency and real-time processing capabilities, requirements that often exceed the practical limits of conventional deep learning models.

\subsection{Binary Classification in Multivariate Time Series}\label{sec:binclass}

We can view a multivariate time series, \(\mathbf{X}\), as a sequence of temporally ordered observations, represented as

\[
\mathbf{X} = \{\mathbf{x}_1, \mathbf{x}_2, \ldots, \mathbf{x}_T\},
\]

where \(\mathbf{x}_t \in \mathbb{R}^n\) is the observation of \(n\) variables at time \(t\), and \(T\) is the total number of time steps in \(\mathbf{X}\). We will discuss two distinct forms of time series binary classification:

% These objectives can be categorized into four distinct cases:

\subsubsection*{Independent Sample Classification}

The classifier ignores any temporal patterns and assigns a binary label to each observation independently. Each prediction is based solely on the current observation \(\mathbf{x}_t\). This can be expressed as:

\[
f: \mathbf{X} \to \{0, 1\},
\]

where \(f(\mathbf{x}_t)\) predicts the binary label for time \(t\).

\subsubsection*{Stateful Step-Wise Classification}\label{sec:statefulformulation}

The classifier assigns a binary label to each time step \(t\), using only the current observation \(\mathbf{x}_t\) along with an internal state \(\mathcal{H}_t\) that encodes information about prior observations (\(\mathbf{X}_{1:t-1}\)). \(\mathcal{H}_t\) represents a compressed internal representation of past observations. This allows for more efficient temporal modeling without explicitly storing the entire sequence. This can be formalized as:

\[
f: \mathbf{X} \times \mathcal{H} \to \{0, 1\},
\]

where the internal state evolves dynamically over time, being updated recursively based on the prior state and observation. In some cases, such as with \acp{RNN}, the state transition is explicitly defined:

\[
\mathcal{H}_t = g(\mathcal{H}_{t-1}, \mathbf{x}_{t-1}),
\]

where \(g\) is the state transition function. In an \acp{SNN}, the state transition function is not explicitly defined but rather emerges from the inherent dynamics of the model, described in more detail in Section~\ref{sec:methodology}.

% The final binary label is then predicted as:

% \[
% y_t = f(\mathbf{x}_t, \mathcal{H}_t).
% \]

\subsection{Neuromorphic Computing}
Neuromorphic systems offer an energy-efficient solution for binary classification in multivariate time-series data. \acp{SNN} are naturally suited to this domain due to their stateful dynamics (neurons maintain internal membrane potentials over time), enabling implicit modeling of temporal dependencies at the individual neuron level, without the need to store full input histories (as in transformers) or define special recurrent structures (as in RNNs). 
Unlike conventional \acp{ANN}, \acp{SNN} encode and process time-varying data through sparse, event-driven spikes. 
This asynchronous computation also minimizes redundant processing, allowing for low-latency, low-power inference suitable for deployment on edge devices with strict energy and memory constraints \cite{schuman2022opportunities,krestinskaya2019neuromemristive,mitchell2020caspian}.

Despite theoretical advantages over traditional \acp{ANN}, \acp{SNN} have not yet gained widespread adoption in the mainstream machine learning community. One major challenge is determining optimal network architectures for specific tasks, particularly when targeting resource-constrained hardware. Unlike in \acp{ANN}, where well-defined architectural standards exist for many applications, the appropriate design of \acp{SNN} is less understood. This challenge is compounded by the need for compact networks that operate efficiently on neuromorphic hardware platforms. While methods like surrogate gradients \cite{neftci2019surrogate} and EventProp \cite{wunderlich2021event} enable the application of gradient descent in specific cases, these approaches require fixed architectures prior to training, which can lead to the loss of some of the potential benefits offered by \acp{SNN}. 

% Gradient-based training algorithms, such as backpropagation, often rely on predefined architectures and make simplifying assumptions that do not fully leverage the unique temporal and spiking dynamics of \acp{SNN}. 
% However, unlike in \acp{ANN}, where extensive research has established well-defined architectural standards optimized for various tasks, the design of \ac{SNN} architectures remains an open challenge. 

Evolutionary algorithms, such as \ac{EONS} \cite{schuman2020evolutionary}, offer an alternative for developing \acp{SNN}. By simultaneously evolving both the architecture and parameters, \ac{EONS} can explore a wide range of network configurations without requiring predefined structures. 
% This is particularly advantageous for edge-based systems, where the optimal trade-off between accuracy, sparsity, and energy efficiency is task-dependent and hardware-constrained. 
Moreover, evolutionary approaches naturally accommodate the event-driven nature of spiking neurons, enabling the creation of compact, efficient networks tailored for deployment on neuromorphic platforms like \microCaspian{} \cite{mitchell2020caspian}.

\subsection{Focus of this Study}

In this work, we leverage \ac{EONS} to design highly compact \acp{SNN} optimized for stateful, step-wise binary classification tasks in multivariate time series. By training these networks directly on hardware simulators within the TENNLab framework, we ensure that the resulting \acp{SNN} are not only efficient but also deployable on energy-constrained edge devices.

We propose a general framework and procedure for training and evaluating \acp{SNN} for binary classification problems in multivariate time series. Our approach inherently yields \acp{SNN} capable of direct deployment on \microCaspian{} \cite{mitchell2020caspian}.
% In our evaluation of this framework, we seek to address two questions: (1) How does the approach generalize 

To evaluate the approach, we first conduct an in-depth study on a challenging low-\ac{SNR} gamma-ray detection problem. We extend prior work \cite{ghawaly2022neuromorphic, ghawaly2023performance} in this domain by introducing improved dataset design and novel training procedures. We then deploy the trained \ac{SNN} on microCaspian{} and perform a power analysis, comparing the results against an alternative low-power solution implemented on a \ac{MCU}.

Building on the insights gained from gamma-ray detection, we extend our methods to an additional application to assess the generalizability of our framework \emph{without} domain-specific optimization: seizure detection in scalp \ac{EEG} recordings.
% We demonstrate our approach by applying it to a challenging application, the detection of low signal-to-noise ratio source signatures in gamma-ray radiation spectra. 
% We extend prior work \cite{ghawaly2022neuromorphic, ghawaly2023performance} in this domain by introducing improved dataset design and novel training procedures. We deploy the trained \ac{SNN} on \microCaspian{} and perform a power analysis, comparing the results against an alternative low-power solution implemented on a \ac{MCU}.

% This paper is structured as follows: 
% \begin{itemize} 
% \item Section~\ref{sec:related_work} reviews related work in neuromorphic computing applied to binary classification in time series
% \item Section~\ref{sec:methodology} describes the hardware-software codesign process followed to develop deployable \acp{SNN} for binary classification problems in multivariate time series data
% \item Section~\ref{sec:applications} outlines the process of applying the methodology to each of the three selected applications, including any specific considerations taken for that application.
% \item Section\ref{sec:results} presents and discusses the results for each application
% \item Section~\ref{sec:discussion} discusses the results more generally, highlighting the key findings and making recommendations for future research to improve the results
% \item Section~\ref{sec:conclusion} concludes the study and outlines plans for future work.
% \end{itemize}

\section{Related Work}\label{sec:related_work}

Neuromorphic computing has garnered increasing interest in the research community for its application to anomaly detection in time-series data. One approach that has emerged is the use of \acp{OESNN}, which identify relationships within input data by capturing temporal patterns from unlabeled normal data and encoding them as the network's normal state. 
% This normal state represents the baseline against which the network evaluates incoming data. 
When the input data  causes the network's state to deviate from the learned normal state, an anomaly is declared. Studies utilizing \acp{OESNN} include the work of \cite{macikag2021unsupervised}, which focuses on univariate input data, and \cite{bassler2022unsupervised}, which extends the approach to multivariate inputs. These methods learn by pruning or adding neurons as new data are encountered, so as to maintain optimal performance.

Another proposed method is the \emph{predict and evaluate} approach \cite{cherdo2023time}. In this method, the \ac{SNN} learns to forecast the next state of the system and then calculates the error between the predicted and actual state. If the error exceeds a predefined threshold, the model declares an anomaly. This approach can be applied to both univariate and multivariate time-series, depending on how the inputs are structured and encoded.

In contrast to these, our work focuses on binary classification using labeled datasets rather than anomaly detection. 

The work most related to ours is \cite{gaurav2023reservoir}, where the authors propose two \ac{SNN} models inspired by \ac{RC} and Legendre Memory Units, to address univariate time series classification. The first, \ac{SLRC}, follows a conventional \ac{RC} design and is deployed on Intel's Loihi platform. The second, \ac{LSNN}, introduces nonlinearity in the readout layer and is trained with Surrogate Gradient Descent (SurrGD). Their models achieve high performance when compared against liquid state machines, however, the authors note that their approach can only be applied to univariate, not multivariate time series. Additionally, our work differs in that we target the \microCaspian FPGA-based neuromorphic device, which consumes less power than Loihi, at the expense of reduced maximum network size.

While hardware efficiency is often cited as a motivating factor, true hardware-software co-design is not explored in any of the reviewed prior work, except for that proposed by \cite{gaurav2023reservoir}. In contrast, we train and evaluate our networks directly in a virtual simulator that replicates the \microCaspian neuromorphic device. We also deploy the trained \ac{SNN} on an actual \microCaspian device and perform a preliminary power analysis.

Finally, all of the reviewed works focus on optimizing a single \ac{SNN} for anomaly detection. As discussed in Section \ref{sec:intro}, the design of effective \ac{SNN} architectures for different applications is poorly understood. Prior work, such as \cite{neculae2020ensembles}, has demonstrated the potential for ensembles of \acp{SNN} to improve performance. While they highlight the benefits of ensemble learning, they do not explore methods to evolve individual network architectures. In contrast, our method trains many unique \acp{SNN} using \ac{EONS} and then ensembles up to 3 \acp{SNN}. 

% By combining offline evolutionary training, comprehensive architecture mutation, an ensemble voting mechanism, and concrete hardware co-design, we aim to present a methodology that diverges from current practices in time-series neuromorphic learning and offers improved adaptability and robustness for anomaly detection.

\section{General Methodology}\label{sec:methodology}

Our proposed methodology develops compact \acp{SNN} optimized for binary classification in multivariate time series data, designed for deployment on the low-power \microCaspian platform. While application-specific considerations are necessary for optimal performance, the framework is intended to be broadly generalizable as starting point for development. It comprises the following steps: dataset preparation, training on virtualized hardware using \ac{EONS} within the TENNLab framework, test set evaluation of the trained network population, network ensembling, final model selection and evaluation on hardware. Methods specific to each of the investigated applications are presented in their respective portions of Section~\ref{sec:applications}.

In our framework, each \ac{SNN} is designed based on the \emph{Stateful Step-wise Binary Classification} paradigm outlined in Section~\ref{sec:statefulformulation}. Specifically, the \ac{SNN} maps the spike-encoded current observation, \(\mathbf{s}_t\), and an internal state encoding prior observations, \(\mathcal{H}_{t-1}\), to a binary class label \(\{0, 1\}\). Here, \(\mathbf{s}_t\) is a train of spikes of value \(1\) occurring at set time intervals. Formally, this mapping can be expressed as:

\[
f: \mathcal{S} \times \mathcal{H} \;\to\; \{0, 1\},
\]

where \(\mathbf{s}_t \in \mathcal{S}\) represents the spike-encoded inputs, and \(\mathcal{H}\) denotes the space of possible internal states.

The internal state at a given time step \(\mathcal{H}_t\) is updated recursively based on the prior state and the previous observation:

\[
\mathcal{H}_t = \Psi(\mathcal{H}_{t-1}, \mathbf{s}_{t-1}),
\]

where \(\Psi\) represents an abstract function of the internal state dynamics of the \ac{SNN}. The SNN begins each inference at \(\mathcal{H}_t\) and processes \(\mathbf{s}_{t}\) through its network of \ac{IF} neurons for a pre-defined inference time, \(\tau\). After running for \(\tau\), the \ac{SNN}'s state consists of the membrane potentials of its neurons and the spiking activity (in-flight after reaching \(\tau\)), which depend on the network's architecture, synaptic weights, neuron thresholds, and previous inputs. Collectively, these processes guide the state transition and encode temporal dependencies, enabling effective processing of sequential data.

For a time series of dimensionality \(n \times T\), each network has \(n\) or \(2n\) input neurons (one for each variable), depending on the encoder binning scheme selected (see Section~\ref{sec:encoders}), and a single output neuron. The total \emph{spike count} of the output neuron, denoted \(z_t\), is a non-negative integer representing the number of spikes emitted by the output neuron over the inference period \(\tau\). We obtain the final binary label \(y_t\) by thresholding:
\[
y_t \;=\;
\begin{cases}
1, & \text{if } z_t > \theta, \\
0, & \text{otherwise},
\end{cases}
\]
where \(\theta\) is a user-defined threshold. Hence, whenever \(z_t\) exceeds \(\theta\), the \ac{SNN} predicts a label of 1; otherwise, it predicts 0. In some applications, it may be beneficial to threshold on a rolling sum of spike count with a fixed time window, to threshold on consecutive bursts of spikes. This is beneficial for applications with a high data sampling rate, where spurious noise-induced spikes could increase \ac{FPR}.

\subsection{Spike Encoding}\label{sec:encoders}

For an observation at a given time step, \(\mathbf{x}_t \in \mathbb{R}^n\), the encoder maps the observation to a corresponding spike train \(\mathbf{s}_t\). Each component \(x_t^{(i)}\) in \(\mathbf{x}_t\) has a minimum (\(\min_i\)) and maximum (\(\max_i\)) value, either derived from the physics of the variable itself or determined empirically from a training dataset that adequately covers the domain of observations. The spike-encoded signal \(\mathbf{s}_t\) thus contains a series of binary pulses over an encoding window of length \(\tau\), which is the inference time for the \ac{SNN}.

In prior work on the application discussed in Section~\ref{sec:neurorad}, a hyperparameter exploration procedure was performed across the variety of encoding schemes offered through the TENNLab framework, with two methods demonstrating the highest success: \emph{rate} and \emph{spikes}. We normalize each variable, \(x_t^{(i)}\), to 
\[
\tilde{x}_t^{(i)} \;=\; \frac{x_t^{(i)} - \min_i}{\max_i \;-\; \min_i}.
\]
    
\begin{itemize}
    \item \emph{Rate Encoding}: Given the encoding window of length \(\tau\), spikes of value \(1\) are distributed at a frequency proportional to \(\tilde{x}_t^{(i)}\). Formally, we model this frequency as 
    \(
    \tilde{x}_t^{(i)} \times \lambda,
    \)
    where \(\lambda\) is the scaling factor indicating the maximum firing rate. In our setting, \(\lambda = \tau\), meaning that a fully normalized input \(\tilde{x}_t^{(i)} = 1\) corresponds to one spike per unit time, yielding \(\tau\) spikes over the entire inference window. Values above \(\max_i\) saturate to  \(\lambda\).
    \item \emph{Spikes Encoding}: In this approach, rather than distributing spikes continuously over the entire inference window, the encoder emits spikes at uniform unit-time intervals. The input amplitude is translated into the total number of spikes generated, ranging from \(0\) up to \(\tau\). 
    % Concretely, for a fully normalized variable \(\tilde{x}_t^{(i)} = 1\), the encoder outputs one spike per time step for the entire window, yielding \(\tau\) spikes in total; smaller values produce fewer spikes proportionally. Values above \(\max_i\) saturate to  \(\tau\).

\end{itemize}

In our implementation of \emph{spikes} encoding, we incorporate an encoder binning strategy, which distributes spikes into \(b\) neurons (or ``bins''), where each bin \(b_i\) handles a specific fractional subrange of \(\min_i\) to \(\max_i\). This binning process can help preserve finer distinctions in the encoded input while retaining a discrete spiking representation. Finally, the binning strategy leverages a feature called ``flip-flop,'' wherein even-numbered bins reverse the order of spikes. In other words, for these bins, the maximum firing rate occurs at the minimum value of the corresponding subrange. This design choice can be beneficial for waveforms that oscillate around zero, providing a more balanced representation of positive and negative values. 

\subsection{\microCaspian{}}\label{sec:ucaspian} % Aaron Young
This work leverages the Caspian\cite{mitchell2020caspian} simulator and the \microCaspian{}\cite{mitchell2020ucaspian} \ac{FPGA}-based \ac{NEP} both developed at \ac{ORNL} and designed to integrate as a processor within the TENNLab Framework.
The Caspian simulator is a parameterizable event-based neuromorphic hardware simulator, and for this work it is configured with parameters to match the \microCaspian{} hardware implementation with equivalent deterministic behavior. 
The \microCaspian{} \ac{NEP} core is the target \ac{SNN} hardware accelerator.
It is built into the embedded neuromorphic platform's FPGA as presented in \cite{witherspoon2024event} and used for edge deployment and power analysis in this work.

% uCaspian Features
\microCaspian{} is specifically designed to fit within the constrained resources of a Lattice iCE40UP5K FPGA. This FPGA is small, low-power, and inexpensive, making it well suited for low-power edge applications. 
\microCaspian{} implements the \ac{IF} neuron model with 8-bit thresholds, 16-bit charge state, and 4-bit axonal delay (0-15 cycles).
Each synapse has an 8-bit signed weight value. 
\microCaspian{} only supports axon delay, without support for synaptic delay. %However, any network using synapse delay can be converted into an equivalent network using only anoxal delay by adding additional intermediate neurons for each unique incoming synapse delay.
Spike routing is handled through target neuron lookup in the synapse routing tables. 
This allows for very flexible routing within the \ac{SNN} network, with support for self-connections, recurrent connections, and up to an all-to-all connection pattern. 
The max size of an \ac{SNN} is limited by the memory used to store the \ac{SNN} configurations, allowing up to 256 neurons and 4096 synapses, with any connection pattern.

% uCaspian Architecture
% The \microCaspian{} \ac{NEP} is implemented as an event-based processing pipeline, with stages that match the components of the SNN model.
% The key components are communication, control, synapses, dendrites, neurons, and axons.
% Each network cycle of operation consists of flushing the charge from the previous cycle's dendrite memory to the neuron stage, the neuron stage applies charge and checks for new spike events, and the spikes are sent to the axon unit to have the configured delay added.
% After the axon delay, the spikes are sent to the control unit, where they are passed through the synapse stage to convert the binary spikes into a charge before being sent to the correct target dendrite to wait for the next network cycle. Each stage along the pipeline is event driven, with each stage running in parallel, operating on events in the processing stream, and each stage must complete all incoming events before the processor is ready to move on to the next cycle. Because of this architecture, the time required to evaluate a network cycle and the power consumed per cycle depends on the activity present in the network during that cycle and varies cycle-to-cycle. The SNN loaded onto the processor, internal state, and input contribute to the current cycle's activity.
The \microCaspian{} \ac{NEP} is an event-driven pipeline that mirrors the SNN model’s stages: communication, control, synapses, dendrites, neurons, and axons. In each cycle, dendrite charges move to neurons, which check for spikes, add any configured axon delays, and pass spikes through synapses back into dendrites. Because all stages run in parallel and only process events, the time and power per cycle vary with network activity, internal states, and inputs.

\subsection{Training}\label{sec:training}
\subsubsection{\ac{EONS}}
In this work, we train \acp{SNN} with the \ac{EONS} approach. As an evolutionary algorithm, \ac{EONS} begins with a population of candidate \acp{SNN}, denoted by
\[
\mathcal{P}^{(g)} \;=\; \bigl\{ S_{1}^{(g)}, \, S_{2}^{(g)}, \,\ldots,\, S_{N}^{(g)} \bigr\},
\]
where \(S_{i}^{(g)}\) represents the  \(i\)-th \ac{SNN} in generation \(g\) and \(N\) is the population size. Each \ac{SNN} \(S_{i}^{(g)}\) is either randomly initialized or initialized using pre-created networks.  These networks are evaluated using the fitness function \(F\) (described below), and each network is assigned a fitness score \(F\bigl(S_{i}^{(g)})\).  These fitness scores are used to inform the selection process, which is used to choose which networks will serve as parents for the next generation. In this work, we use \emph{tournament selection}: a random subset of individuals,

\[
\mathcal{T} \;=\; \bigl\{ S_{j_1}^{(g)}, \ldots, S_{j_k}^{(g)} \bigr\} \;\subseteq\; \mathcal{P}^{(g)},
\]

compete in a tournament, and then the best individual in that subset is selected as the parent. Formally, if \(\mathcal{T}\) contains \(k\) individuals, then the parent chosen from \(\mathcal{T}\) is
\[
\widetilde{S}^{(g)} \;=\; \arg\max_{\,S \,\in\, \mathcal{T}} \, F(S).
\]

This process allows for a nice tradeoff between preferentially selecting better performing networks while still allowing for the chance that lower performing individuals are also occasionally selected (i.e., in the case that the tournament is entirely made up of lower performing individuals). 

Once parents are selected, reproduction operations are applied to generate the next generation \(\mathcal{P}^{(g+1)}\). \ac{EONS} uses three different reproduction operators: 
\begin{itemize}
    \item \emph{Duplication (Cloning)}: A straightforward copy of a parent \(\widetilde{S}^{(g)}\) into the next generation.
    \item \emph{Crossover}: Given two parents, \(\widetilde{S}_{p_1}^{(g)}\) and \(\widetilde{S}_{p_2}^{(g)}\), the offspring inherit both structural and parameter characteristics from each parent.
    \item \emph{Mutation}: With some probability, a parent's architecture or parameters are altered. Possible mutations include adding or removing neurons/synapses or perturbing parameters such as synaptic weights or neuronal thresholds.
\end{itemize}

\ac{EONS} also uses elitism which guarantees that some number of the best performing individuals from \(\mathcal{P}^{(g)}\) are cloned into \(\mathcal{P}^{(g+1)}\). Once the children have been created, they replace the parent population and are evaluated using the fitness function. From there, this process of selection, reproduction, and evaluation continues over the course of a pre-specified number of generations or a fixed amount of time. The final population after training is completed is denoted as \(\mathcal{P}^{(*)}\).

Table~\ref{tab:eons-hparams} contains the values of the EONS hyperparameters used for all of the applications discussed in Section~\ref{sec:applications}. The meaning of each of these parameters is outlined in the EONS documentation and prior work \cite{schuman2020evolutionary}.

\begin{table}[t!]
    \centering
    \caption{Selected EONS Hyperparameters within Tennlab}
    \label{tab:eons-hparams}
    \begin{tabular}{l l}
        \hline
        \textbf{Parameter} & \textbf{Value} \\
        \hline
        \texttt{starting\_nodes}              & 50 \\
        \texttt{starting\_edges}              & 50 \\
        \texttt{merge\_rate}                  & 0 \\
        \texttt{population\_size}             & 100 \\
        \texttt{multi\_edges}                 & 0 \\
        \texttt{crossover\_rate}              & 0.5 \\
        \texttt{mutation\_rate}               & 0.9 \\
        \texttt{tournament\_size\_factor}     & 0.1 \\
        \texttt{tournament\_best\_net\_factor}& 0.9 \\
        \texttt{random\_factor}               & 0.05 \\
        \texttt{num\_mutations}               & 4 \\
        \texttt{node\_mutations}              & \{\texttt{Threshold}: 1.0\} \\
        \texttt{edge\_mutations}              & \{\texttt{Weight}: 0.7, \texttt{Delay}: 0.3\} \\
        \texttt{num\_best}                    & 3 \\
        \texttt{add\_node\_rate}              & 0.5 \\
        \texttt{delete\_node\_rate}           & 0.25 \\
        \texttt{add\_edge\_rate}              & 0.75 \\
        \texttt{delete\_edge\_rate}           & 0.25 \\
        \texttt{node\_params\_rate}           & 2.5 \\
        \texttt{edge\_params\_rate}           & 2.5 \\
        \hline
    \end{tabular}
\end{table}

When training an \ac{SNN} through \ac{EONS}, our objective is to optimize both \(\Psi\) and \(f\) to maximize a fitness function \(F\). This involves two key components:
\begin{itemize}
    \item Evolving an architecture and the weights that enable \(\Psi\) to effectively encode salient historical features in the model's internal state \(\mathcal{H}\), maintaining short-term memory of relevant information.
    \item Developing a function \(f\) that utilizes this short-term memory (\(\mathcal{H}\)) and the current observation (\(\mathbf{x}_t\)) to classify the input as either \(0\) or \(1\).
\end{itemize}

During training, the threshold \(\theta\) is set to 0, to encourage the \ac{SNN} to only produce output spikes during class-1 encounters. This differs in evaluation, as described in Section~\ref{sec:eval_methods}.

\subsubsection{Fitness Function Selection}
\label{sec:fitness-function-selection}

The choice of \(F\) plays a critical role in training. With EONS, the developer has the freedom to design a fitness function without the restrictions of differentiability imposed by training algorithms based on gradient descent. As such, metrics based directly on the discrete binary performance, such as classification accuracy, precision, recall, or F1-score, can be directly incorporated into the fitness function. 

For most applications, we have found that the \ac{MCC} performs well as a fitness function. The \ac{MCC} is a metric that measures the correlation between the true and predicted binary classes and considers all four components of the binary confusion matrix. It is robust to unbalanced datasets and is regarded as one of the most informative measures of a binary classifier's performance~\cite{chicco2020advantages, chicco2021matthews, boughorbel2017optimal}. The MCC is defined as

% \begin{equation}
% \label{eq:mcc}
\[
\text{MCC} = \frac{TP \cdot TN - FP \cdot FN}{\sqrt{(TP + FP)(TP + FN)(TN + FP)(TN + FN)}}
\]
% \end{equation}

% In this equation, \( TP \), \( TN \), \( FP \), and \( FN \) represent the true positives, true negatives, false positives, and false negatives, respectively. 
and ranges from \(-1\) (perfect inverse correlation) to \(+1\) (perfect correlation), with \(0\) indicating no correlation.

% In applications with specific requirements, such as that described in Section~\ref{sec:neurorad}, where performance at very low false positive rates are critical, \ac{MCC} may not be the best choice. 

Metrics that are sensitive to imbalanced data, such as the \ac{AUC}, were found to be poor choices for fitness functions. \acp{SNN} trained for \ac{AUC} tended to converge to a local minimum of predicting the majority class. Data balancing procedures may mitigate this issue; however, one could simply select metrics like the \ac{MCC}, that are inherently robust to imbalance.

\subsection{Evaluation}\label{sec:eval_methods}

One of the benefits of \ac{EONS} is that repeated training procedures can produce a population of diverse candidate networks, many of which may be suitable for deployment. From our testing, we have found that it is best to train more than one population of networks for a given application. Depending on the application and computing resources available, anywhere from 100 to 1000 populations are trained. The resulting \acp{SNN} from these population form their own trained population, denoted by
\(\mathcal{P}^{(*)} = \{S_{1}^{(*)}, \dots, S_{N}^{(*)}\}.\) Each \ac{SNN} in \(\mathcal{P}^{(*)}\) is evaluated on the training set using a variety of standard metrics (e.g., accuracy, precision, recall, \ac{MCC}). All possible values of the threshold, \(\theta\), are evaluated, and the one resulting in the maximum \ac{MCC} is then used for sorting the \acp{SNN} in \(\mathcal{P}^{(*)}\). The top 10\% of these are then considered for network ensembling.

\subsection{Network Ensembling}\label{Network-Ensembling}
We have observed that the top \acp{SNN} often have different strengths and weaknesses. We employ \emph{network ensembling} through simple voting procedures to leverage each model's strengths. Specifically, once we identify the top networks based on their training-set \ac{MCC} scores, we evaluate all pairs and trios drawn from these top performers. The ensemble with the highest training set score is then evaluated on the test set. 

For any given input, the ensemble's prediction is obtained via either:
\begin{itemize}
    \item \emph{Any Vote}: The ensemble outputs a class-1 prediction if \emph{any} of the networks predict class 1.
    \item \emph{Majority Vote}: The ensemble outputs class 1 if at least two out of the three networks predict class 1.
\end{itemize}
These voting mechanisms allow us to capitalize on complementary decision boundaries learned by different networks, often resulting in improved overall performance. While more sophisticated ensembling approaches are available, simple voting procedures have the benefit of maintaining interpretability and minimizing computational complexity. They also provide developers the freedom to operate the ensemble members either serially or in parallel. 

% It is important to emphasize that the ensemble classification is based on single-network decisions, rather than thresholding on spikes. As such, one must optimize \(\theta\) individually for each network before casting votes. The determination of the optimal threshold often requires application-specific considerations. In Section~\ref{sec:applications}, unless stated otherwise, the threshold selected is the one that maximizes \ac{MCC} on the training set.

% \subsection{Model Selection}

% For final model selection, we recommend performing an additional evaluation of the top networks and their ensembles on test data collected directly in the field. In our work, we only reached this stage for a single application, discussed in Section~\ref{sec:neurorad}. Even if a model performs well on a curated test set, there is no guarantee of equally strong performance under real-world conditions (cf. Section~\ref{sec:discussion}). For many applications, it is infeasible to create a test set that fully spans the phase space of all relevant variables. Consequently, networks that generalize well from training to standard testing may still falter once additional field tests introduce novel conditions or data distributions not captured in previous evaluations.

\section{Applications}\label{sec:applications}

\subsection{Source Detection in Gamma-ray Spectral Data}\label{sec:neurorad}
\subsubsection{Background}

In this application, we train \acp{SNN} to detect radioactive sources in gamma-ray spectral time series data. We define a \emph{source} as any radioactive object or material that is not part of the natural radioactive \emph{background}. The intensity and energy distribution (spectrum) of gamma radiation measured by a sensor can vary significantly due to differences in the concentrations of naturally occurring radioactive nuclides. Consequently, the background signal is highly variable in both magnitude and shape, which makes identifying anomalous signals (i.e., potential sources) more challenging.

Our study uses a spectroscopic detector, specifically a NaI(Tl) crystal measuring 3"\(\times\)3" in volume, which records the energy deposited by incoming gamma-rays. Each source of interest exhibits a unique set of gamma-ray energies, and these can be exploited for both detection and identification. However, the intensity of the gamma-ray signature decreases according to an inverse square law (\(1/r^2\)) with respect to the distance \(r\) between source and detector, and any intervening shielding can further attenuate the signal. 

This combination of factors introduces significant variability into the measured spectra, resulting in a low \ac{SNR}. In prior work, detection algorithms built on \acp{ANN} have demonstrated strong performance in this application\cite{anderson2021radiation, ghawaly2022arad}; however, their energy consumption and latency can limit their viability for long-term edge-based operation.

\subsubsection{Dataset}

Building on the experimentally validated framework used for generating synthetic datasets in prior work~\cite{ghawaly2020data, nicholson2020generation}, we created a new synthetic dataset featuring additional sources and greater background variation. This dataset simulates a 3\("\times"\)3~NaI(Tl) detector moving at walking speed through a virtual model of the 7000-area at \ac{ORNL}. 
% Each material in the model is assigned a distribution of KUT concentrations that can be sampled to produce different background conditions. 
We simulate 20 unique source/shielding combinations, each at 8 distinct \ac{SNR} levels. Here, \(\text{SNR} = \frac{S}{\sqrt{S + B}}\), where \(S\) is the total source counts and \(B\) is the total background counts measured during an encounter. To generate individual runs, we sample from the background distributions and simulate the detector passing each source at various speeds and standoff distances.

We also collected over 8 hours of real-world dynamic background radiation spectra by driving a detector around the Knoxville, Tennessee region and recording data. Half of these data were included in training and half in  testing. Synthetic sources were also injected over random sub-ranges of the measured background runs. Table~\ref{tab:neurorad-dataset} below provides a breakdown of the datasets.

\begin{table}[ht]
    \centering
    \caption{Dataset Parameters for the Gamma-ray Source Detection Application}
    \label{tab:neurorad-dataset}
    \begin{tabularx}{\linewidth}{l X X}
    \hline
    \textbf{Parameter} & \textbf{Training} & \textbf{Test} \\
    \hline
    \multicolumn{3}{l}{\emph{Spectral Integration}} \\
    \hline
    Integration Time 
      & 0.5\,s 
      & \texttt{-} \\
    Integration Stride 
      & 0.5\,s
      & \texttt{-} \\
    Calibration Range 
      & 1\,keV--3001\,keV 
      & \texttt{-} \\
    Number of Energy Bins 
      & 32 
      & \texttt{-} \\
    Binning Structure 
      & $\sqrt{E}$ 
      & \texttt{-} \\
    \hline
    \emph{Background Variability} 
      & 80\% about nominal\cite{ghawaly2020data} 
      & \texttt{-} \\
    \hline
    \multicolumn{3}{l}{\emph{Sources}} \\
    \hline
    SNM Sources 
      & 25\,kg HEU, 8\,kg WGPu
      & \texttt{-} \\
    Point Sources 
      & $^{241}$Am, $^{133}$Ba, $^{57}$Co, $^{60}$Co, \\
      & $^{137}$Cs, $^{177}$Lu
      & \texttt{-} \\
    SNM Shielding 
      & bare, steel, Cu, Pb
      & \texttt{-} \\
    Point Source Shielding 
      & bare, steel
      & \texttt{-} \\
    Source SNR 
      & 8 steps up to 16
      & 8 steps up to 18 \\  % <-- Different here
    \hline
    \multicolumn{3}{l}{\emph{Speed \& Offset}} \\
    \hline
    Relative Speed (Synthetic) 
      & 0.5\,--1.5\,m/s
      & 0.5\,--2.2\,m/s \\  % <-- Different here
    Detector/Source Offset (Synthetic)
      & 2\,m, 4\,m
      & \texttt{-} \\
    Relative Speed (Injected) 
      & 2.2\,--13.4\,m/s
      & \texttt{-} \\
    \hline
    \multicolumn{3}{l}{\emph{Dataset Size}} \\
    \hline
    Background-only 
      & 240 runs
      & \texttt{-} \\
    Containing Source 
      & 3520 runs
      & \texttt{-} \\
    \hline
    \end{tabularx}
\end{table}

\subsubsection{Data Preprocessing}

Each run in the dataset consists of a time series of gamma-ray spectra, with each spectrum represented by 32 energy bins. The integer value in each bin corresponds to the number of gamma rays detected in that bin’s energy range at a given time step. Prior to spike encoding, each spectrum is normalized to an integral sum of one. We then generate spike trains using \emph{rate} encoding, as described in Section~\ref{sec:methodology}. This method was chosen because it performed slightly better than \emph{spikes} encoding for this application~\cite{ghawaly2022neuromorphic}.

\subsubsection{Training}
We train 1000 populations of 100 networks using the procedure described in Section~\ref{sec:methodology}. This application demands a very low \ac{FPR}, consistent with standards such as ANSI~N42.53~\cite{ansin42532013} and ANSI~N42.35~\cite{ansin42352016}, which specify maximum \acp{FAR} of \(1\,\text{hr}^{-1}\) and \(0.5\,\text{hr}^{-1}\), respectively. The International Atomic Energy Agency further recommends a maximum \ac{FAR} of \(1/12\,\text{hr}^{-1}\) \cite{iaea2004detection}.

To meet these requirements, we define a metric, denoted \(\text{TPR}_0\), which is the true positive rate of an \ac{SNN} at a \ac{FAR} of \(0\,\text{hr}^{-1}\). Our fitness function is then:
% \begin{equation}
% \label{eq:fitness_rad}
\[
F \;=\; \text{F1} \;+\; \text{TPR}_0^2
\]
% \end{equation}
where \(\text{F1}\) is the harmonic mean of precision and recall. This metric is inspired by that used in \cite{ghawaly2023performance}. After each training epoch, \ac{ROC} curves are generated for each network, from which \(\text{TPR}_0\) is derived.

Initially, we attempted using \(F = \text{TPR}_0\) alone, but during early training stages, most networks could not detect any sources at a \ac{FAR} of \(0\,\text{hr}^{-1}\). As a result, \(\text{TPR}_0\) was zero for nearly all networks, causing the training process to stagnate. Incorporating \(\text{F1}\) into the objective enables the networks to reach a non-zero \(\text{TPR}_0\) and thus effectively optimize both the overall detection performance and the low \ac{FAR} requirement.

Due to the large size of our training set, each epoch uses a training batch that is a random 1\% subset of the whole. We also implement a new feature, referred to as \emph{SNR scaling}, where we gradually increase the difficulty of this training batch for each epoch. We do this by biasing samples toward high SNR early in training and sampling uniformly in later epochs.

% At epoch \( t \), the training batch sampling distribution is determined by a weight factor:
% \[
% \alpha = 1 - \frac{\min(t, k)}{k},
% \]
% where \( k \) is the epoch at which the distribution becomes uniform, and in this case, is set to 600, which is the number of epochs/generation we train using \ac{EONS}.

% Each sample in the training set is weighted by its peak \ac{SNR}, scaled by \( \alpha \):
% \[
% w_i = (\text{SNR}_i)^\alpha, \quad \forall i \in \{1, \dots, N\}.
% \]
% The weights are normalized to probabilities:
% \[
% p_i = \frac{w_i}{\sum_{j=1}^N w_j}.
% \]

% Using these probabilities, \( n \) indices are sampled with replacement, transitioning from high-SNR sampling in early epochs to uniform sampling as \( t \to k \).

After the 1000 populations were trained and evaluated on the test set, the top 10 networks were selected for ensembling evaluation. The optimal thresholds for these networks were set to yield an ensemble \ac{FAR}$\leq 1 \text{hr}^{-1}$.

\subsubsection{Comparisons}

We compare our results to two well-established algorithms, the \ac{SAD}~\cite{sad1, sad2} and the \ac{ARAD}~\cite{ghawaly2022arad}. \ac{SAD} is a simple but effective source detection algorithm that learns a low-dimensionality linear representation of background spectra (32 energy bins) using \ac{PCA}, using spectral reconstruction error to detect non-background sources. \ac{ARAD} operates in a similar manner to \ac{SAD}, but utilizes a very small ($\sim$3k parameters) deep convolutional autoencoder to learn a nonlinear representation of background radiation spectra (128 energy bins). Both the \ac{ARAD} and \ac{SAD} algorithms perform binary classification under the \emph{Independent Sample Classification} paradigm outlined in Section~\ref{sec:intro}. Operationally, \ac{SAD} provides the best alternative to \acp{SNN} in a low-power setting, as it can easily be implemented on a \ac{MCU}, as described in Section~\ref{sec:power}. \ac{ARAD} has been shown to outperform \ac{SAD} in highly dynamic background source detection scenarios when utilizing large NaI(Tl) detectors~\cite{ghawaly2022arad}.

\begin{figure}[!t]
\centering
\includegraphics[width=0.5\textwidth]{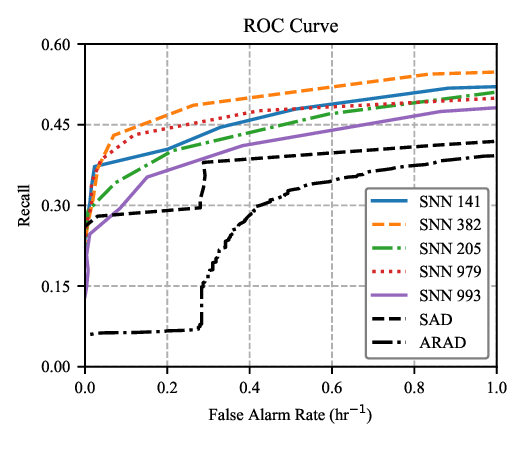}
\caption{Testing dataset ROC curves for the top 5 \acp{SNN} and reference \ac{ARAD} and \ac{SAD} algorithms.}
\label{fig:neurorad_roc}
\end{figure}
\subsection{Seizure Detection in EEG}\label{sec:seizure}

We also evaluate our approach is seizure detection in \acf{EEG} data. We use the CHB-MIT Scalp EEG database~\cite{guttag2010chbmit,shoeb2009application,goldberger2000components}, a publicly available dataset of pediatric \ac{EEG} recordings collected at Children’s Hospital Boston. It comprises 686 continuous EEG records from 23 pediatric subjects with intractable seizures, divided into 24 cases. Each case contains between 9 and 42 continuous \texttt{.edf} records. The dataset includes 198 annotated seizures and details the onset and duration of each event. 

These recordings include high-resolution signals sampled at 256~Hz, with most containing 23 \ac{EEG} channels following the International 10-20 electrode placement system. The data is continuously recorded~\cite{wong2023eeg} for durations of 1, 2, and 4-hours. 

Unlike our in-depth analysis of gamma-ray spectral data, the intent behind the examination of this dataset is to evaluate the performance of our general methodology on a new application without implementing any domain-specific optimizations.

% Like the LENDB analysis, the examination of this dataset is preliminary, with the primary objective of testing the generalization of our methodology without domain-specific optimizations.

\subsubsection{Data Preprocessing}
\begin{table}[t!]
\centering
\begin{tabularx}{\linewidth}{@{}l p{0.46\linewidth} p{0.12\linewidth} p{0.12\linewidth}@{}}
\toprule
\textbf{Dataset} & \textbf{Patients} & \textbf{Seizure Records} & \textbf{Total} \\ 
\midrule
\textbf{Training Set} & chb12, chb15, chb04, chb01, chb05, chb03, chb02, chb18, chb13 & 100 & 368 \\
\textbf{Validation Set} & chb11, chb06, chb10 & 18 & 80 \\
\textbf{Test Set} & chb21, chb20, chb22, chb19, chb14, chb24, chb16, chb17, chb08, chb09, chb07, chb23 & 71 & 294 \\
\bottomrule
\end{tabularx}

\vspace{1em}
\caption{Dataset splits across patients and records.}
\label{tab:eeg_data}
\end{table}

To ensure consistency across records, data from 18 sensors were used as the \ac{SNN} input. Records containing multiple seizures were divided into separate runs, each featuring a single seizure. This resulted in a dataset composed of 189 seizures across 742 records, with 2 being excluded due to missing or inconsistent annotations.

The records were trimmed to a maximum duration of 400 seconds, fully encapsulating the seizure events, keeping an amount of padding at the beginning and end of the seizure. This padding was chosen randomly between 8 and 20~s to introduce variability in the start times of the seizure events. This trimming procedure was performed to fully capture the temporal dynamics of seizure events while minimizing record length for computational efficiency. 
Records with a duration shorter than 400 seconds were unmodified. 

The dataset was then divided across patients and records into training (50\%), validation (10\%), and testing (40\%) subsets. The test set is fully comprised of patients not seen in training or validation. Table~\ref{tab:eeg_data} presents the final dataset split statistics.

\subsubsection{Training}

Training follows the general procedure in Section~\ref{sec:methodology}. To account for the high sampling rate, we apply a 20~s rolling sum over the spike output during testing (evaluation only), with the detection threshold applied to this sum. This window length was chosen empirically to encompass seizure duration. The rolling sum is omitted during training to better encourage the network to suppress false positives.

\subsubsection{Comparisons}
Several alternative approaches have been evaluated on the CHB-MIT Scalp EEG dataset, often incorporating specialized preprocessing or model architectures that yield a range of performance outcomes. For instance, \cite{bhattacharya2022automatic} employs a fully connected \ac{ANN} feeding into an \ac{LSTM}, achieving a sensitivity of 97.746 and a \ac{FPR} of 0.2373 across 21 patients. Similarly, \cite{fergus2015automatic} apply frequency-based techniques to further refine feature extraction, achieving a reported sensitivity of 93.

\begin{figure}[t!]
    \centering
    \includegraphics[width=0.5\textwidth]{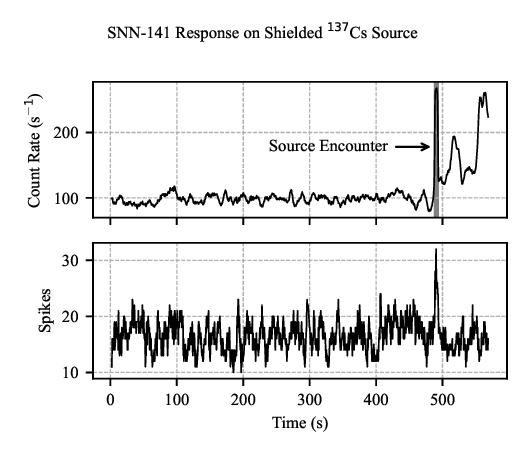}
    \caption{Spike response of \emph{\ac{SNN} 141} on a time series from the testing dataset, with a synthetic shielded $^{137}$Cs source injected on measured background.}
    \label{fig:snn141_fig}
\end{figure}
\section{Results and Discussion}\label{sec:results}

% This section presents and discusses the results for each application.

\subsection{Source Detection in Gamma-ray Spectral Data}\label{sec:results-neurorad}

\figurename \ref{fig:neurorad_roc} shows the \ac{ROC} curves generated on the full testing dataset, for the top 5 networks, as defined by their TPR at a \ac{FAR} of 1hr$^{-1}$ on the training set. The ROC curve is shown for \acp{FAR} from 0 to 2hr$^{-1}$, displaying the full range of relevance based on United States and international recommendations~\cite{ansin42342015, ansin42352016, ansin42532013}. The ROC curves for the reference \ac{SAD} and \ac{ARAD} algorithms are also shown.

The top performing \ac{SNN} was \emph{\ac{SNN} 141}, with TPR$_1=0.498$. \emph{\ac{SNN} 141} comprises 49 neurons and 66 synapses, with a highly recurrent architecture. \figurename \ref{fig:snn141_fig} shows the spike response of \emph{\ac{SNN} 141} on a run from the testing set containing a synthetic shielded $^{137}$Cs source injected over real-world  background. On the test set, TPR$_1=0.518$ for \emph{\ac{SNN} 141}.

All unique ensembles of the top-100 networks from training were evaluated and sorted according to their TPR$_1$ scores on the training set. The top-780 ensembles were all trio ensembles, with the highest performing one reaching TPR$_1=0.671$, a 29.5\% improvement. The highest performing pair ensemble reached TPR$_1=0.647$, a 24.9\% improvement over \emph{\ac{SNN} 141}. All ensembles that outperformed \emph{\ac{SNN} 141} used the \emph{any} voting method. \figurename \ref{fig:neurorad_ensemble} shows the performance of the top  ensemble compared against its individual constituents, \emph{\ac{SNN} 135}, \emph{\ac{SNN} 33}, and \emph{\ac{SNN} 75}, on the non-shielded sources in the test set.

\subsubsection{Discussion} As described in \cite{ghawaly2023performance}, many of the sources are simulated at very low \acp{SNR}, thus very high performance on this dataset may not be achievable. The results should be interpreted relative to the reference algorithm performance, which have been established in the community.

% As seen on \figurename \ref{fig:neurorad_roc}, all of the top-10 trained \acp{SNN} outperformed both reference algorithms. Both \ac{SAD} and \ac{ARAD} detect non-background sources by thresholding the spectral reconstruction error between the input spectrum and a learned background reconstruction, a method that has proven effective for larger detectors (e.g. the 2"$\times$4"$\times$16" NaI(Tl) as used in \cite{ghawaly2020data, ghawaly2022arad}. However, our dataset was collected/simulated with a significantly smaller detector (see Section~\ref{sec:neurorad}). Because gamma-ray detection follows a Poisson process, the uncertainty in each energy bin scales with $\sqrt{N}$, where $N$ is the number of counts in said bin. Thus, for a given integration time, smaller detectors yield fewer overall counts and correspondingly higher relative noise. In this study, we suspect that the trained \acp{SNN} outperformed \ac{SAD} and \ac{ARAD} in part due to improved robustness to noise. We also hypothesize the ability for the \acp{SNN} to identify both spectral and temporal patterns contributed to this, confirming the findings of \cite{ghawaly2023performance}, which showed that short-term memory improved \ac{SNN} performance in this application.

As shown in \figurename~\ref{fig:neurorad_roc}, the top five \acp{SNN} outperformed both \ac{SAD} and \ac{ARAD}, which work well on larger detectors but are less robust for smaller ones. Smaller detectors yield fewer counts, increasing relative noise from Poisson statistics. We hypothesize that the \acp{SNN}’ ability to model both spectral and temporal patterns improves noise robustness and temporal pattern recognition, aligning with \cite{ghawaly2023performance}, which highlighted the benefits of short-term memory in this domain.

\figurename~\ref{fig:snn141_fig} illustrates the spiking behavior of the top-performing \ac{SNN} under a variety of conditions: from 0 to about 480\,s, the background remains relatively steady with no source present; around 480--490\,s, a shielded \({}^{137}\mathrm{Cs}\) source is introduced; after 490\,s, only background is observed, but it becomes highly dynamic, likely due to the detector passing buildings. As shown, the \ac{SNN} exhibits low output neuron spiking activity during both steady and dynamic background conditions, while sharply increasing with source presence.

\begin{figure}[!t]
\centering
\includegraphics[width=0.5\textwidth]{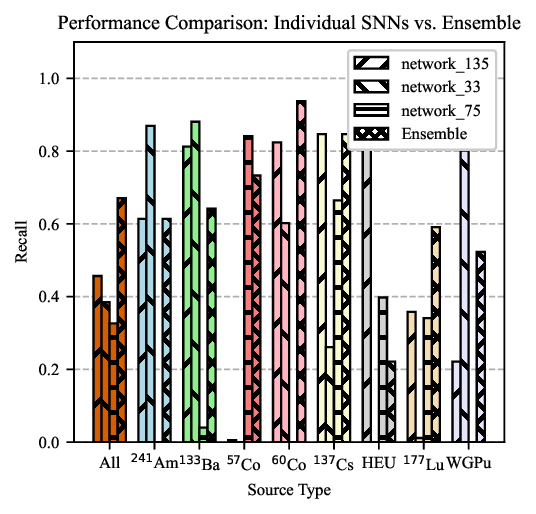}
\caption{Recall for the top ensemble and its individual constituents for all non-shielded sources and all sources together (All).}
\label{fig:neurorad_ensemble}
\end{figure}

Finally, ensembling either two or three of the top \acp{SNN} boosted performance by more than 15\% over single-network results. \figurename \ref{fig:neurorad_ensemble} demonstrates how \emph{any} vote ensembling allows the ensemble to leverage the collective strengths of each of the constituent \acp{SNN}, with different networks performing better than others on different source types.
% Notably, the top peforming single \ac{SNN}, \\emph{\ac{SNN} 141}, is not part of the top performing ensemble. 

In the case of \microCaspian{} where each network in the ensemble is run sequentially, ensembling would increase latency and energy consumption roughly proportional to the number of networks in the ensemble. This trade-off may be viable, however, for applications that can tolerate additional overhead. 
%Additionally, if the networks are small enough, they can likely be run in parallel with a sub-linear scaling of latency and energy consumption.
% Additionally, if the networks are small enough they can be run as a single combined network, or they can be run in parallel by including multiple \microCaspian cores, providing trade-offs between latency and energy consumption.

\subsection{Power Analysis}\label{sec:power}

A power analysis was conducted to compare our SNN implementation with the \ac{SAD} algorithm on a custom embedded platform that integrates an STM32U5 \ac{MCU} and a Lattice iCE40 UltraPlus \ac{FPGA}. The \microCaspian{} \ac{SNN} core resides in the \ac{FPGA}, while the \ac{MCU} handles spike encoding and decoding. This custom hardware platform was designed with the algorithmic methods described herein, and is described in greater detail in prior work~\cite{witherspoon2024event}.

The \ac{SAD} algorithm runs entirely on the \ac{MCU} using the ARM CMSIS-DSP library with single-precision floating-point arithmetic. Power measurements were performed with a Keysight N6705C analyzer, bypassing on-board regulators to power the 1.2V (FPGA core) and 3.3V (I/O and \ac{MCU}) rails separately. When idle, the \ac{MCU} consumes about 6mW, and the \ac{FPGA} clock can be disabled to 300µW. During inference, \ac{SAD} executes in roughly 100µs (plus idle-transition overhead) at 160MHz, requiring an estimated 27~µJ per inference.

In contrast, the SNN consumes about 2mW at 24MHz for 20.2ms during inference, or 40.4µJ, plus an additional 32mW (646.4µJ) while transferring spikes between the \ac{MCU} and \ac{FPGA}. This overhead occurs because software-based encoding keeps the \ac{MCU} from entering low-power states.

\subsubsection{Discussion} These results demonstrate that power management is a complex co-design issue: both the \ac{MCU} and \ac{FPGA} remain in high-power states during inference. Future improvements involve reducing the \ac{MCU} clock, refining the communication protocol, and leveraging DMA or lower-power communication to reduce data-movement overhead. 
% Further hardware integration, granular power domains, and more efficient sensor interfacing could also help address these challenges. 
Finally, while this system currently supports hot-swapping of \acp{SNN} for serial ensemble member operation, future work will support parallel operation of multiple \acp{SNN} with shared inputs.

\subsection{Seizure Detection in EEG}\label{subsec:seizure-detection-results}
% \subsubsection{Discussion}
The testing set results are presented in Table~\ref{tab:eeg_results}. The highest performing single model achieved an \ac{MCC} score of 0.47 and was comprised of 18 neurons connected by 23 synapses with high recurrence. Ensembling proved to be highly effective for enhancing network performance, with trio ensembles achieving the highest scores. Performance changes in \ac{MCC} scores were observed as follows: +2.13\% for the pair ensemble, -34.04\% for the trio any-vote ensemble, and +51.06\% for the trio majority-vote ensemble. 

The pair ensemble demonstrated only a modest gain in performance. The any-vote method for the trio ensemble suffered a decline in \ac{MCC} and an increase in \ac{FPR}. In contrast, the majority-vote ensemble yielded significantly higher performance, achieving the best overall performance with a TPR of 0.95 and an FPR of 0.16.

The majority-vote trio ensemble results are competitive with prior work on the CHB-MIT EEG dataset. For instance, \cite{bhattacharya2022automatic} reported a TPR of 0.97 and an FPR of 0.24, while \cite{fergus2015automatic} achieved a TPR of 0.93 and an FPR of 0.06. However, both approaches relied on extensive preprocessing and artifact extraction, whereas our method achieved comparable performance using raw data.

These results demonsrate the generalizability of our methodology with minimal application-specific modifications. For maximum performance on specific applications, one should consider domain-specific design choices in fitness function and data preprocessing/encoding. This is standard practice for any \ac{AI} application and can be automated using hyperparameter optimization tools. 
% We recommend Bayesian search~\cite{snoek2012practical} or tools like the Data Efficient Framework for Exploration (Deffe)~\cite{liu2020deffe} to improve hyperparameter optimization efficiency.

\begin{table}[t!]
\centering
% Reduce intercolumn spacing locally
\setlength{\tabcolsep}{4pt} 
\begin{tabular}{@{}p{1cm} c c c c c c c c@{}}
\toprule
\textbf{Model} & \textbf{Ensemble Method} & \textbf{TP} & \textbf{TN} & \textbf{FP} & \textbf{FN} & \textbf{TPR} & \textbf{FPR} & \textbf{MCC} \\
\midrule
Individual & -- & 27 & 212 & 8 & 47 & 0.36 & 0.04 & 0.47 \\
Pair  & \textit{vote} & 25 & 216 & 4 & 49 & 0.33 & 0.02 & .48 \\ 
Trio & \textit{any vote} & 32 & 167 & 53 & 42 & 0.43 & 0.24 & 0.31 \\
Trio & \textit{majority vote} & 70 & 184 & 36 & 4 & 0.95 & 0.16 & \textbf{0.71} \\
\bottomrule
\end{tabular}
\vspace{1em}
\caption{EEG testing set performance metrics for the top individual SNN, top pair ensemble, and top trio ensemble.}
\label{tab:eeg_results}
\end{table}

% Using network-242
% MCU single core 160 MHz with FPU
% FPGA uCaspian core 25 MHz
% SAD on STM32U5 @ 160MHz : 
% SAD 100us 55mW peak / 1 ms total @ 27mW average \approx 27 uJ
% SNN on ice40 1.0V : 2mW * 20.2ms \approx 40.4 uJ w/ MCU 32mW * 20.2ms = 646.4 uJ

% \section{General Discussion}\label{sec:discussion}

% The successful application of \ac{SNN}-based classifiers to the LENDB dataset underscores the versatility of our approach. Despite the inherent differences between seismic waveforms and gamma-ray spectral data, the \ac{SNN} framework maintained robust performance, highlighting its potential for cross-domain applications. However, it is important to note that the analysis conducted on the LENDB dataset did not delve as deeply into optimization and fine-tuning as performed for the gamma-ray dataset. This more preliminary evaluation serves to demonstrate the foundational applicability of our general framework, paving the way for more in-depth studies and potential optimizations tailored to seismic data in future work.

\section{Conclusion}\label{sec:conclusion}

In this work, we proposed a general-framework for training and evaluating \acp{SNN} for binary classification in multivariate time series. We performed a thorough validation of this framework, with application-specific optimizations, on a challenging dataset involving the detection of low \ac{SNR} radioactive sources, comparing against alternative deep learning and \ac{PCA}-based approaches. We then tested the generalizability of the approach without adapting it for specific applications, by evaluating it on seizure detection in EEG waveforms. Based on our results, we make the following conclusions:

\begin{enumerate}
    \item The proposed framework is an effective solution for binary classification tasks in multivariate time series, with the resulting networks being directly deployable on the \microCaspian{} neuromorphic hardware platform.
    \item With application-specific optimizations on data preprocessing, spike encoder selection, and fitness function selection, the framework can yield \acp{SNN} with performance competitive with or exceeding alternative low-resource algorithms. This is evident in the results on the radiation dataset, where the \ac{SNN} outperformed both the \ac{PCA} and deep learning approach.
    \item \ac{EONS} offers the benefit of evolving \acp{SNN} with diverse capabilities, allowing one to leverage simple voting ensembling across two or three \acp{SNN} to yield significant performance gains.
    \item The low-power \microCaspian{} neuromorphic computing platform is a good choice for this domain, performing inference on \(\sim\)2~mW of power. 
\end{enumerate}

\section*{Acknowledgment}
This research used resources of the Experimental Computing Laboratory (ExCL) at the Oak Ridge National Laboratory, which is supported by the Office of Science of the U.S. Department of Energy under Contract No. DE-AC05-00OR22725. 

Portions of this research were conducted with high performance computational resources provided by Louisiana State University (http://www.hpc.lsu.edu) and also the Louisiana Optical Network Infrastructure (http://www.loni.org).
% \textit{ANONYMIZED}

% \section*{References}

\bibliographystyle{IEEEtran} 
\bibliography{references}

\begin{thebibliography}{10}
\providecommand{\url}[1]{#1}
\csname url@samestyle\endcsname
\providecommand{\newblock}{\relax}
\providecommand{\bibinfo}[2]{#2}
\providecommand{\BIBentrySTDinterwordspacing}{\spaceskip=0pt\relax}
\providecommand{\BIBentryALTinterwordstretchfactor}{4}
\providecommand{\BIBentryALTinterwordspacing}{\spaceskip=\fontdimen2\font plus
\BIBentryALTinterwordstretchfactor\fontdimen3\font minus \fontdimen4\font\relax}
\providecommand{\BIBforeignlanguage}[2]{{%
\expandafter\ifx\csname l@#1\endcsname\relax
\typeout{** WARNING: IEEEtran.bst: No hyphenation pattern has been}%
\typeout{** loaded for the language `#1'. Using the pattern for}%
\typeout{** the default language instead.}%
\else
\language=\csname l@#1\endcsname
\fi
#2}}
\providecommand{\BIBdecl}{\relax}
\BIBdecl

\bibitem{strubell2020energy}
E.~Strubell, A.~Ganesh, and A.~McCallum, ``Energy and policy considerations for modern deep learning research,'' in \emph{Proceedings of the AAAI conference on artificial intelligence}, vol.~34, no.~09, 2020, pp. 13\,693--13\,696.

\bibitem{villalobos2024position}
\BIBentryALTinterwordspacing
P.~Villalobos, A.~Ho, J.~Sevilla, T.~Besiroglu, L.~Heim, and M.~Hobbhahn, ``Position: Will we run out of data? limits of {LLM} scaling based on human-generated data,'' in \emph{Forty-first International Conference on Machine Learning}, 2024. [Online]. Available: \url{https://openreview.net/forum?id=ViZcgDQjyG}
\BIBentrySTDinterwordspacing

\bibitem{valmeekam2022large}
K.~Valmeekam, A.~Olmo, S.~Sreedharan, and S.~Kambhampati, ``Large language models still can't plan (a benchmark for llms on planning and reasoning about change),'' in \emph{NeurIPS 2022 Foundation Models for Decision Making Workshop}, 2022.

\bibitem{mirzadeh2024gsm}
I.~Mirzadeh, K.~Alizadeh, H.~Shahrokhi, O.~Tuzel, S.~Bengio, and M.~Farajtabar, ``Gsm-symbolic: Understanding the limitations of mathematical reasoning in large language models,'' \emph{arXiv preprint arXiv:2410.05229}, 2024, unpublished.

\bibitem{jiang2024peek}
B.~Jiang, Y.~Xie, Z.~Hao, X.~Wang, T.~Mallick, W.~J. Su, C.~J. Taylor, and D.~Roth, ``A peek into token bias: Large language models are not yet genuine reasoners,'' \emph{arXiv preprint arXiv:2406.11050}, 2024, in press.

\bibitem{roy2019towards}
K.~Roy, A.~Jaiswal, and P.~Panda, ``Towards spike-based machine intelligence with neuromorphic computing,'' \emph{Nature}, vol. 575, no. 7784, pp. 607--617, 2019.

\bibitem{schuman2022opportunities}
C.~D. Schuman, S.~R. Kulkarni, M.~Parsa, J.~P. Mitchell, P.~Date, and B.~Kay, ``Opportunities for neuromorphic computing algorithms and applications,'' \emph{Nature Computational Science}, vol.~2, no.~1, pp. 10--19, 2022.

\bibitem{calimera2013human}
A.~Calimera, E.~Macii, and M.~Poncino, ``The human brain project and neuromorphic computing,'' \emph{Functional neurology}, vol.~28, no.~3, p. 191, 2013.

\bibitem{schuman2017survey}
C.~D. Schuman, T.~E. Potok, R.~M. Patton, J.~D. Birdwell, M.~E. Dean, G.~S. Rose, and J.~S. Plank, ``A survey of neuromorphic computing and neural networks in hardware,'' \emph{arXiv preprint arXiv:1705.06963}, 2017, unpublished.

\bibitem{krestinskaya2019neuromemristive}
O.~Krestinskaya, A.~P. James, and L.~O. Chua, ``Neuromemristive circuits for edge computing: A review,'' \emph{IEEE transactions on neural networks and learning systems}, vol.~31, no.~1, pp. 4--23, 2019.

\bibitem{mitchell2020caspian}
\BIBentryALTinterwordspacing
J.~P. Mitchell, C.~D. Schuman, R.~M. Patton, and T.~E. Potok, ``Caspian: A neuromorphic development platform,'' in \emph{Proceedings of the Neuro-Inspired Computational Elements Workshop}, ser. NICE '20.\hskip 1em plus 0.5em minus 0.4em\relax New York, NY, USA: Association for Computing Machinery, 2020. [Online]. Available: \url{https://doi.org/10.1145/3381755.3381764}
\BIBentrySTDinterwordspacing

\bibitem{neftci2019surrogate}
E.~O. Neftci, H.~Mostafa, and F.~Zenke, ``Surrogate gradient learning in spiking neural networks: Bringing the power of gradient-based optimization to spiking neural networks,'' \emph{IEEE Signal Processing Magazine}, vol.~36, no.~6, pp. 51--63, 2019.

\bibitem{wunderlich2021event}
T.~C. Wunderlich and C.~Pehle, ``Event-based backpropagation can compute exact gradients for spiking neural networks,'' \emph{Scientific Reports}, vol.~11, no.~1, p. 12829, 2021.

\bibitem{schuman2020evolutionary}
C.~D. Schuman, J.~P. Mitchell, R.~M. Patton, T.~E. Potok, and J.~S. Plank, ``Evolutionary optimization for neuromorphic systems,'' in \emph{Proceedings of the 2020 Annual Neuro-Inspired Computational Elements Workshop}, 2020, pp. 1--9.

\bibitem{ghawaly2022neuromorphic}
J.~Ghawaly, A.~Young, D.~Archer, N.~Prins, B.~Witherspoon, and C.~Schuman, ``A neuromorphic algorithm for radiation anomaly detection,'' in \emph{Proceedings of the International Conference on Neuromorphic Systems 2022}, 2022, pp. 1--6.

\bibitem{ghawaly2023performance}
J.~Ghawaly, A.~Young, A.~Nicholson, B.~Witherspoon, N.~Prins, M.~Swinney, C.~Celik, C.~Schuman, and K.~Patel, ``Performance optimization study of the neuromorphic radiation anomaly detector,'' in \emph{Proceedings of the 2023 International Conference on Neuromorphic Systems}, 2023, pp. 1--7.

\bibitem{macikag2021unsupervised}
P.~S. Maci{\k{a}}g, M.~Kryszkiewicz, R.~Bembenik, J.~L. Lobo, and J.~Del~Ser, ``Unsupervised anomaly detection in stream data with online evolving spiking neural networks,'' \emph{Neural Networks}, vol. 139, pp. 118--139, 2021.

\bibitem{bassler2022unsupervised}
D.~B{\"a}{\ss}ler, T.~Kortus, and G.~G{\"u}hring, ``Unsupervised anomaly detection in multivariate time series with online evolving spiking neural networks,'' \emph{Machine Learning}, vol. 111, no.~4, pp. 1377--1408, 2022.

\bibitem{cherdo2023time}
Y.~Cherdo, B.~Miramond, and A.~Pegatoquet, ``Time series prediction and anomaly detection with recurrent spiking neural networks,'' in \emph{2023 International Joint Conference on Neural Networks (IJCNN)}.\hskip 1em plus 0.5em minus 0.4em\relax IEEE, 2023, pp. 1--10.

\bibitem{gaurav2023reservoir}
R.~Gaurav, T.~C. Stewart, and Y.~Yi, ``Reservoir based spiking models for univariate time series classification,'' \emph{Frontiers in Computational Neuroscience}, vol.~17, p. 1148284, 2023.

\bibitem{neculae2020ensembles}
G.~Neculae, O.~Rhodes, and G.~Brown, ``Ensembles of spiking neural networks,'' \emph{arXiv preprint arXiv:2010.14619}, 2020.

\bibitem{mitchell2020ucaspian}
\BIBentryALTinterwordspacing
J.~P. Mitchell, C.~D. Schuman, and T.~E. Potok, ``A small, low cost event-driven architecture for spiking neural networks on fpgas,'' in \emph{International Conference on Neuromorphic Systems 2020}, ser. ICONS 2020.\hskip 1em plus 0.5em minus 0.4em\relax New York, NY, USA: Association for Computing Machinery, 2020. [Online]. Available: \url{https://doi.org/10.1145/3407197.3407216}
\BIBentrySTDinterwordspacing

\bibitem{witherspoon2024event}
B.~Witherspoon and A.~Young, ``Event-driven sensing and embedded neuromorphic platforms for gamma radiation monitoring,'' in \emph{Proceedings of the Great Lakes Symposium on VLSI 2024}, 2024, pp. 779--784.

\bibitem{chicco2020advantages}
D.~Chicco and G.~Jurman, ``The advantages of the matthews correlation coefficient (mcc) over f1 score and accuracy in binary classification evaluation,'' \emph{BMC genomics}, vol.~21, pp. 1--13, 2020.

\bibitem{chicco2021matthews}
D.~Chicco, N.~T{\"o}tsch, and G.~Jurman, ``The matthews correlation coefficient (mcc) is more reliable than balanced accuracy, bookmaker informedness, and markedness in two-class confusion matrix evaluation,'' \emph{BioData mining}, vol.~14, pp. 1--22, 2021.

\bibitem{boughorbel2017optimal}
S.~Boughorbel, F.~Jarray, and M.~El-Anbari, ``Optimal classifier for imbalanced data using matthews correlation coefficient metric,'' \emph{PloS one}, vol.~12, no.~6, p. e0177678, 2017.

\bibitem{anderson2021radiation}
C.~M. Anderson-Cook, D.~Archer, M.~S. Bandstra, J.~C. Curtis, J.~M. Ghawaly, T.~H. Joshi, K.~L. Myers, A.~D. Nicholson, and B.~J. Quiter, ``Radiation detection data competition report,'' Los Alamos National Laboratory (LANL), Los Alamos, NM (United States), Tech. Rep., 2021.

\bibitem{ghawaly2022arad}
J.~M. Ghawaly~Jr, A.~D. Nicholson, D.~E. Archer, M.~J. Willis, I.~Garishvili, B.~Longmire, A.~J. Rowe, I.~R. Stewart, and M.~T. Cook, ``Characterization of the autoencoder radiation anomaly detection (arad) model,'' \emph{Engineering Applications of Artificial Intelligence}, vol. 111, p. 104761, 2022.

\bibitem{ghawaly2020data}
J.~M. Ghawaly, A.~D. Nicholson, D.~E. Peplow, C.~M. Anderson-Cook, K.~L. Myers, D.~E. Archer, M.~J. Willis, and B.~J. Quiter, ``Data for training and testing radiation detection algorithms in an urban environment,'' \emph{Scientific data}, vol.~7, no.~1, pp. 1--6, 2020.

\bibitem{nicholson2020generation}
A.~D. Nicholson, D.~E. Peplow, J.~M. Ghawaly, M.~J. Willis, and D.~E. Archer, ``Generation of synthetic data for a radiation detection algorithm competition,'' \emph{IEEE Transactions on Nuclear Science}, vol.~67, no.~8, pp. 1968--1975, 2020.

\bibitem{ansin42532013}
``American national standard performance criteria for backpack-based radiation-detection systems used for homeland security,'' \emph{ANSI N42.53-2013}, pp. 1--48, 2013.

\bibitem{ansin42352016}
``American national standard for evaluation and performance of radiation detection portal monitors for use in homeland security,'' \emph{ANSI N42.35-2016 (Revision of ANSI N42.35-2006)}, pp. 1--70, 2016.

\bibitem{iaea2004detection}
``Detection of radioactive materials at borders,'' International Atomic Energy Agency, Tech. Rep., 2002.

\bibitem{sad1}
K.~Miller and A.~Dubrawski, ``Gamma-ray source detection with small sensors,'' \emph{IEEE Transactions on Nuclear Science}, vol.~65, no.~4, pp. 1047--1058, 2018.

\bibitem{sad2}
P.~Tandon, P.~Huggins, R.~Maclachlan, A.~Dubrawski, K.~Nelson, and S.~Labov, ``Detection of radioactive sources in urban scenes using bayesian aggregation of data from mobile spectrometers,'' \emph{Information Systems}, vol.~57, pp. 195--206, 2016.

\bibitem{guttag2010chbmit}
\BIBentryALTinterwordspacing
J.~Guttag, ``{CHB-MIT Scalp EEG Database (version 1.0.0)},'' PhysioNet, 2010. [Online]. Available: \url{https://doi.org/10.13026/C2K01R}
\BIBentrySTDinterwordspacing

\bibitem{shoeb2009application}
A.~Shoeb, ``Application of machine learning to epileptic seizure onset detection and treatment,'' Ph.D. dissertation, Massachusetts Institute of Technology, September 2009.

\bibitem{goldberger2000components}
A.~Goldberger, L.~Amaral, L.~Glass, J.~Hausdorff, P.~C. Ivanov, R.~Mark, and H.~E. Stanley, ``Physiobank, physiotoolkit, and physionet: Components of a new research resource for complex physiologic signals,'' \emph{Circulation}, vol. 101, no.~23, pp. e215--e220, 2000, [Online].

\bibitem{wong2023eeg}
S.~Wong, A.~Simmons, J.~Rivera-Villicana, S.~Barnett, S.~Sivathamboo, P.~Perucca, Z.~Ge, P.~Kwan, L.~Kuhlmann, R.~Vasa, K.~Mouzakis, and T.~J. O'Brien, ``{EEG datasets for seizure detection and prediction—A review},'' \emph{Epilepsia Open}, vol.~8, no.~2, pp. 252--267, 2023.

\bibitem{bhattacharya2022automatic}
A.~Bhattacharya, ``Automatic seizure prediction using cnn and lstm,'' \emph{arXiv preprint arXiv:2211.02679}, 2022.

\bibitem{fergus2015automatic}
P.~Fergus, D.~Hignett, A.~Hussain, D.~Al-Jumeily, and K.~Abdel-Aziz, ``Automatic epileptic seizure detection using scalp eeg and advanced artificial intelligence techniques,'' \emph{BioMed research international}, vol. 2015, no.~1, p. 986736, 2015.

\bibitem{ansin42342015}
``American national standard performance criteria for handheld instruments for the detection and identification of radionuclides,'' \emph{ANSI N42.34-2015 (Revision of ANSI N42.34-2006)}, pp. 1--60, 2016.

\end{thebibliography}

\end{document}